\documentclass[preprint,12pt]{elsarticle} 
\usepackage[T1]{fontenc}
\usepackage[utf8]{inputenc}
\usepackage{lmodern}
\usepackage{amsmath,amssymb}
\usepackage{siunitx}         
\usepackage{graphicx}
\usepackage{booktabs}        
\usepackage{threeparttable}  
\usepackage{hyperref}        
\usepackage{enumitem}
\usepackage{geometry}
\usepackage{xcolor}
\usepackage{hyperref}
\usepackage{xurl}                
\hypersetup{hidelinks,breaklinks=true}
\usepackage{tabularx}
\usepackage{booktabs,threeparttable}
\newcolumntype{Y}{>{\raggedright\arraybackslash}X} 
\usepackage{subcaption}
\usepackage[protrusion=true,expansion=true]{microtype} 
\biboptions{numbers,sort&compress}

\journal{Journal of Diabetes and Its Complications}

\title{Generalizable Diabetes Risk Stratification via Hybrid Machine Learning Models}

\begin{document}
\begin{frontmatter}

\author[1]{Athar Parvez}
\ead{g202393830@kfupm.edu.sa}

\author[1]{Muhammad Jawad Mufti\corref{cor1}}
\ead{g202392310@kfupm.edu.sa}

\address[1]{Information and Computer Science Department, 
King Fahd University of Petroleum and Minerals, 
Dhahran 31261, Saudi Arabia}

\cortext[cor1]{Corresponding author.}

\nonumnote{\textit{Abbreviations and acronyms}:
DM, Diabetes mellitus; NCDs, Non-communicable diseases; IDF, International Diabetes
Federation; ML, Machine Learning; SVM, Support Vector Machine; LR, Logistic Regression;
RF, Random Forest; XGB, Extreme Gradient Boosting (XGBoost); ROC, Receiver Operating
Characteristic; AUROC, Area under the ROC curve; PR, Precision–Recall; AUPRC, Area under
the Precision–Recall curve; SMOTE, Synthetic Minority Over-sampling Technique; RBF,
Radial Basis Function; OGTT, Oral Glucose Tolerance Test; BMI, Body Mass Index; PIMA,
Pima Indians Diabetes Database; EHR, Electronic Health Records; ANN, Artificial Neural
Network; KNN, k-Nearest Neighbors; DBN, Deep Belief Network; CGRU, Convolutional Gated
Recurrent Unit; GAN, Generative Adversarial Network; NAS, Neural Architecture Search;
TRIPOD-AI, Transparent Reporting of a multivariable prediction model for Individual
Prognosis Or Diagnosis—AI extension.}

\begin{abstract}
\textbf{Background/Purpose:}
Diabetes affects over 537 million people worldwide and is projected to reach 783 million by 2045. Early risk stratification can benefit from machine learning. We compare two hybrid classifiers and assess their \emph{generalizability} on an external cohort.

\textbf{Methods:}
Two hybrids were built: (i) XGBoost + Random Forest (XGB--RF) and (ii) Support Vector Machine + Logistic Regression (SVM--LR). A leakage-safe, standardized pipeline (encoding, imputation, min--max scaling; SMOTE on training folds only; probability calibration for SVM) was fit on the primary dataset and \emph{frozen}. Evaluation prioritized \emph{threshold-independent} discrimination (AUROC/AUPRC) and calibration (Brier, slope/intercept). External validation used the PIMA cohort (N=768) with the frozen pipeline; any thresholded metrics on PIMA were computed at the default rule \(\tau=0.5\).

\textbf{Results:}
On the primary dataset (PR baseline = 0.50), XGB--RF achieved AUROC \(\approx 0.995\) and AUPRC \(\approx 0.998\), outperforming SVM--LR (AUROC \(\approx 0.978\); AUPRC \(\approx 0.947\)). On PIMA (PR baseline \(\approx 0.349\)), XGB--RF retained strong performance (AUROC \(\approx 0.990\); AUPRC \(\approx 0.959\)); SVM--LR was lower (AUROC \(\approx 0.963\); AUPRC \(\approx 0.875\)). Thresholded metrics on PIMA at \(\tau=0.5\) were: XGB--RF (Accuracy 0.960; Precision 0.941; Recall 0.944; F1 0.942) and SVM--LR (Accuracy 0.900; Precision 0.855; Recall 0.858; F1 0.857).

\textbf{Conclusions:}
Across internal and external cohorts, XGB--RF consistently dominated SVM--LR and exhibited smaller external attenuation on ROC/PR with acceptable calibration. These results support gradient-boosting–based hybridization as a robust, transferable approach for diabetes risk stratification and motivate prospective, multi-site validation with deployment-time threshold selection based on clinical trade-offs.
\end{abstract}


\end{frontmatter}

\section{Introduction}\label{sec:sec1}
Diabetes mellitus (DM) is a metabolic dis-ease that is chronic in duration and entails im-paired insulin secretion or action with resultant persistent hyperglycemia \cite{ref1}. Diabetes has been described as a global health emergency and con-tributes to a high percentage of the burden from non-communicable diseases (NCDs) and affects millions of individuals worldwide \cite{ref2}. The Inter-national Diabetes Federation (IDF) states that in 2021, approximately 537 million adults (20–79 years) had diabetes and estimated that the numbers would be 783 million in 2045 \cite{ref3}. Diabetes ranks among the top 10 causes of death world-wide and raises the risk of cardiovascular disease, renal failure, and neuropathies, and increases the cost and socioeconomic burden of care \cite{ref4}.\\

There are three primary types of diabetes: type 1, type 2, and gestational diabetes. Type 1 diabetes is an autoimmune condition in which pancreatic beta cells get destroyed and causing total insulin deficiency \cite{ref5}. Type 2 diabetes, on the other hand, is primarily caused due to lifestyle factors including poor dietary habits, excess body fat, inactivity, and genetic factors and thus ac-counts for the largest number of diabetes cases \cite{ref6}. Type 2 diabetes leads to the body developing insulin resistance and the pancreas not being able to produce enough insulin to meet the demand and thus leads to dysregulated glucose metabolism \cite{ref7}. Gestational diabetes mellitus (GDM) occurs in pregnant females due to hormone imbalance and places both the fetus and the mother at risk of developing future metabolic complications \cite{ref8}.\\

The usual signs and symptoms of diabetes are polyuria, polydipsia, polyphagia, fatigue, and blurred vision, and long-term complications in-clude retinopathy, nephropathy, and neuropathy \cite{ref9}. Various factors, such as genetic disposition, high blood pressure, high cholesterol levels, and chronic inflammation, have a significant role in the development of diabetes \cite{ref10}.\\

The advent of machine learning (ML) algorithms has revolutionized predictive analysis in diabetes detection and prognosis in a significant manner. Deep learning, ensemble learning, and hybrid algorithms in ML models enhance diagnostic accuracy through the identification of intrinsic patterns in biomedical data \cite{ref11}. Recent work emphasizes the performance of hybrid ML models in improving prediction accuracy, reducing misclassification rates, and enhancing risk assessment strategies \cite{ref12}. Decision support systems based on artificial intelligence aid clinicians in early detection, personalized treatment, and precision medicine strategies, dramatically improving diabetes management and intervention plans \cite{ref13}. The objectives of the study are as follows:

\begin{itemize}
  \item Compare and evaluate two hybrid machine learning models
        (XGBoost--RF and SVM--LR) for early-stage diabetes prediction
        using standard metrics (Accuracy, Precision, Recall, F1, AUC--PR).
  \item Identify the best-performing hybrid model to support diabetes
        risk stratification and timely clinical intervention.
  \item Assess \textit{generalizability} on an independent cohort (PIMA) by applying the \emph{frozen} preprocessing and models (no refitting, no resampling, no threshold tuning). Prioritize threshold–independent discrimination (AUROC, AUPRC; PR baseline = cohort prevalence) and calibration (Brier, slope/intercept); for completeness, report thresholded metrics at a fixed default rule ($\tau=0.5$). Compare these results against the primary dataset to quantify generalization.

\end{itemize}

The remaining paper's structure is as follows: Section~\ref{sec:sec2} includes related work. In Section~\ref{sec:sec3}, the research material and method are explained. The study's results section is presented in Section~\ref{sec:sec4}. In Section~\ref{sec:sec5}, the conclusion of the research is explained. The last Section~\ref{sec:sec6} is about the future work.

\section{Related Work}\label{sec:sec2}

Hybrid models have also been used for diabetes prediction, utilizing deep learning, ensemble learning, and feature selection among other techniques. A hybrid model consisting of a Stacked 
Autoencoder, Softmax Classifier, and a Genetic Algorithm reported an accuracy of 99.07\%, showcasing the impact of deep learning optimizations. Feature selection was implemented via a Genetic Algorithm (GA) with SVM, Random Forest, KNN, and Naïve Bayes classifiers used in diabetes classification. These classifiers were combined into an ensemble model, which achieved an accuracy of 93.82\%. A different study suggested a hybrid approach combining ensemble methods with deep learning techniques, achieving an accuracy of 80\% \cite{ref14,ref15,ref16}.

A Type-II diabetes diagnosis using a hybrid ensemble-based expert system has also been reported \cite{ref12}. The combination of Artificial Neural Networks, Support Vector Machine, k-Nearest Neighbors, and Naïve Bayes excelled with an accuracy of 98.60\%. This technique outperformed single classifiers evaluated with tenfold cross-validation. In addition, a graphical user interface-based diagnostic tool was proposed for improved clinical usability \cite{ref17}. In another technique based on hybridization-based machine learning, multiple algorithm combinations were tested, where the hybrid of KNN and SVM components was able to obtain the 
highest accuracy of 85\% \cite{ref18}. Using deep learning methods for diabetes prediction and severity grading has led to an accuracy of 99\% when developing the hybrid Convolutional Gated Recurrent Unit (CGRU) model. By leveraging both temporal and spatial features for classification, this model was shown to outperform popular ensemble-based approaches \cite{ref19}.

In addition, a new hybrid framework combined Artificial Neural Networks (ANN) with Genetic Algorithms to optimize regularization techniques for increased predictive reliability. The model achieved an accuracy of 80\%, demonstrating that classifier models are adaptable to different 
datasets and data skewness \cite{ref20}. Another study proposed a hybrid model based on KNN and Logistic Regression for the classification of diabetes, achieving up to 80\% accuracy \cite{ref18}. 

Moreover, a hybrid model based on a Random Forest classification algorithm compared different approaches and found the best performance to be 97.6\% when combining Random Forest with k-means clustering \cite{ref21}. A subsequent study proposed a more advanced deep learning architecture, combining Deep Belief Networks (DBN) with an attention mechanism for diabetes risk prediction. This approach showed progressive improvement through techniques such as Generative Adversarial Networks (GANs) and a hybrid loss function. Among five developed models, the final one reached the best performance with a precision of 0.98, recall of 0.95, F1-score of 0.97, and AUC of 1.00 \cite{ref22}.

While extensive research has been conducted on hybrid machine learning models for diabetes prediction, most studies have focused on demonstrating the superiority of hybrid models over individual (pure) classifiers. These works often aim to maximize predictive accuracy through complex model integrations. However, there is a noticeable gap in the literature when it comes to systematically comparing different hybrid model combinations with each other. This study addresses that gap by conducting a comparative analysis of two distinct hybrid models --- SVM + Naïve Bayes and Random Forest + XGBoost --- to evaluate their relative effectiveness using standard evaluation metrics.

\section{Materials and Methods}\label{sec:sec3}
\begin{figure}[ht] \centering 
\includegraphics[width=0.6\linewidth]{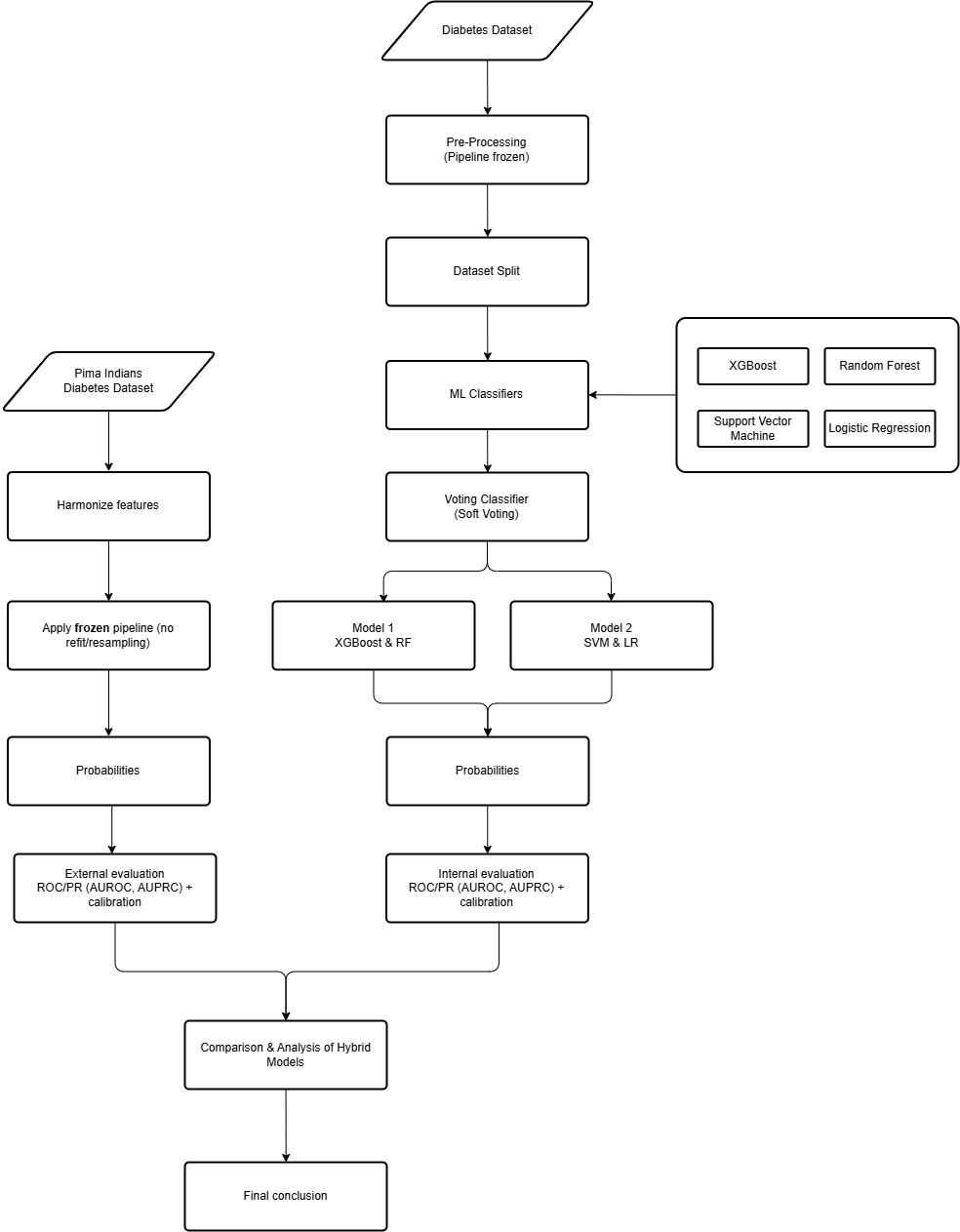} \caption{Research methodology}
\label{fig:methodology}
\end{figure}

To the best of our knowledge, this study is the first to conduct a systematic comparative analysis of hybrid machine learning models for diabetes prediction. A structured methodology (Figure~\ref{fig:methodology}) is adopted to achieve the research goals. The diabetes dataset undergoes extensive pre-processing, including the treatment of missing values, mitigation of class imbalance, transformation of categorical variables, and normalization of the data. To ensure compatibility with machine learning algorithms and enhance predictive performance, categorical features are encoded into numerical representations. Following the pre-processing phase, the dataset is partitioned into training and testing subsets, with 70\% allocated for model training and 30\% reserved for performance evaluation. Hybrid models are then constructed utilizing ensemble-based learning strategies. Specifically, two voting classifier-based hybrid frameworks are developed: one combining Support Vector Machine (SVM) with Logistic Regression (LR), and the other integrating Extreme Gradient Boosting (XGBoost) with Random Forest (RF). These configurations are designed to harness the complementary strengths of individual algorithms. Subsequently, the constructed models undergo rigorous performance evaluation using a suite of standard classification metrics. The study culminates in a detailed comparative analysis to assess the predictive accuracy and overall classification efficacy of the proposed hybrid models.

\subsection{Data Acquisition}
The primary dataset used for experimentation and evaluation was obtained from Kaggle
(\url{https://www.kaggle.com/datasets/iammustafatz/diabetes-prediction-dataset}) and
contains medical and demographic records with a binary diabetes status (positive/negative).
This dataset comprises 100{,}000 patient records and nine attributes in total (eight risk
variables plus the diabetes label). Table~\ref{tab:dataset} provides detailed descriptions
of all variables.

\begin{table}[ht]
\centering
\begin{threeparttable}
\caption{Primary diabetes dataset description}
\label{tab:dataset}
\begin{tabularx}{\linewidth}{@{}llY@{}}
\toprule
\textbf{Feature} & \textbf{Variable Types} & \textbf{Level of Quantization} \\
\midrule
Gender          & Categorical & 3 (female: 0, male: 1, other: 2) \\
Age             & Float       & Age in days \\
Hypertension    & Binary      & 2 (No: 0, Yes: 1) \\
Heart Disease   & Binary      & 2 (No: 0, Yes: 1) \\
Smoking History & Categorical & 5 (No Info: 0, current: 1, ever: 2, former: 3, never: 4, not current: 5) \\
BMI             & Float       & kg/m\textsuperscript{2} \\
HBA1C Level     & Float       & Percentage (\%) \\
Blood Glucose Level & Integer & mg/dL \\
Diabetes        & Binary      & 2 (No: 0, Yes: 1) \\
\bottomrule
\end{tabularx}
\end{threeparttable}
\end{table}

For external validation, we additionally used the PIMA Indians Diabetes Database (UCI
repository; Kaggle mirror: \url{https://www.kaggle.com/datasets/uciml/pima-indians-diabetes-database?resource=download}).
This dataset includes 768 records with eight predictors
(\textit{Pregnancies}, \textit{Glucose}, \textit{BloodPressure}, \textit{SkinThickness},
\textit{Insulin}, \textit{BMI}, \textit{DiabetesPedigreeFunction}, \textit{Age}) and a
binary outcome (\textit{Outcome} $\in \{0,1\}$). A concise description of these variables
appears in Table~\ref{tab:pima-dataset}. Both datasets were processed using the same
pipeline (preprocessing and evaluation metrics) to enable a fair comparison and to assess
the generalizability of the proposed models.

\begin{table}[ht]
\centering
\begin{threeparttable}
\caption{PIMA Indians diabetes dataset description}
\label{tab:pima-dataset}
\begin{tabularx}{\linewidth}{@{}llY@{}}
\toprule
\textbf{Feature} & \textbf{Variable Types} & \textbf{Level of Quantization / Units} \\
\midrule
Pregnancies & Integer (count) & Number of prior pregnancies (0, 1, 2, \dots) \\
Glucose & Integer / Float & 2-hr plasma glucose (OGTT), \textit{mg/dL} \\
BloodPressure & Integer / Float & Diastolic blood pressure, \textit{mmHg} \\
SkinThickness & Integer / Float & Triceps skinfold thickness, \textit{mm} \\
Insulin & Integer / Float & 2-hr serum insulin, \textit{$\mu$U/mL} \\
BMI & Float & Body Mass Index, \textit{kg/m\textsuperscript{2}} \\
DiabetesPedigreeFunction & Float & Diabetes pedigree function (unitless index) \\
Age & Integer & Age in years \\
Outcome & Binary & 2 (No: 0, Yes: 1) \\
\bottomrule
\end{tabularx}
\end{threeparttable}
\end{table}

\subsection{Data Pre-processing}
We applied a consistent pipeline and strictly prevented information leakage by fitting all preprocessing steps on the \emph{primary} training split and then applying these \emph{frozen} transformations to held-out splits and to the external dataset.

\subsubsection{Primary Diabetes Dataset}
Categorical variables were numerically encoded (binary flags for hypertension, heart disease, and diabetes; multi-level encoding for smoking history and gender), with diabetes coded as $1$ (positive) and $0$ (negative). No missing values were detected across the nine features. Continuous variables were scaled via min--max normalization to $[0,1]$. The class distribution was highly imbalanced (non-diabetic $n{=}91{,}500$ vs.\ diabetic $n{=}8{,}500$), see Figure~\ref{fig:primary_distribution}. To mitigate bias during model fitting, we applied the Synthetic Minority Over-sampling Technique (SMOTE) \emph{within training folds only}; validation/test folds remained untouched to avoid leakage. All preprocessing parameters (e.g., encoder mappings, imputation values if needed, and scaler min/max) were estimated on the training data and then applied to the corresponding validation/test splits.

\subsubsection{External Validation PIMA Dataset}
\begin{figure}[ht]
  \centering
  \begin{subfigure}{0.48\linewidth}
    \centering
    \includegraphics[width=\linewidth]{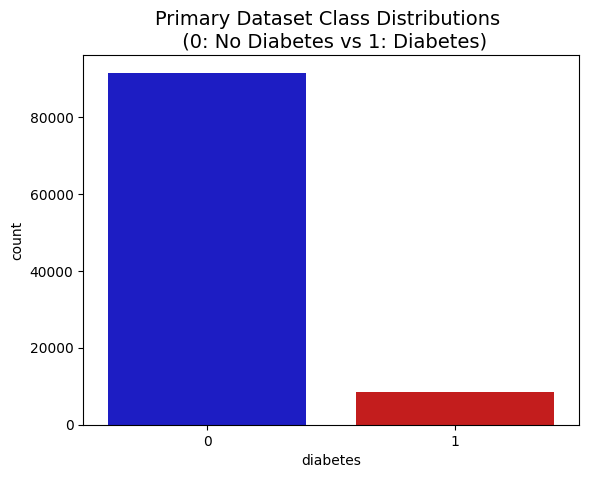}
    \subcaption{Primary dataset class distribution}
    \label{fig:primary_distribution}
  \end{subfigure}\hfill
  \begin{subfigure}{0.48\linewidth}
    \centering
    \includegraphics[width=\linewidth]{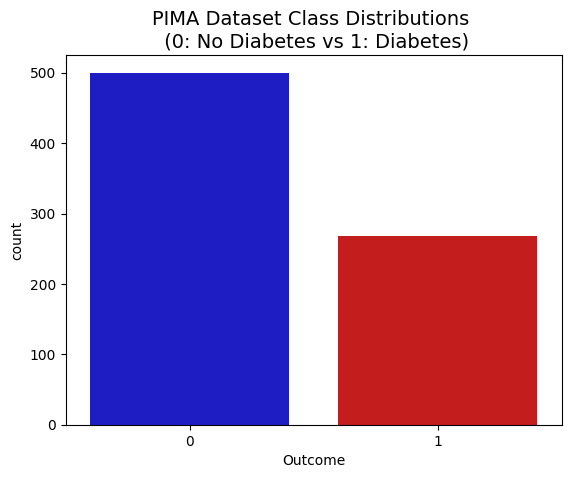}
    \subcaption{PIMA class distribution (0 = No, 1 = Yes)}
    \label{fig:pima_distribution}
  \end{subfigure}
  \caption{Primary and external (PIMA) dataset class distribution.}
  \label{fig:methodology_pima_sidebyside}
\end{figure}

For external evaluation, we used the PIMA Indians Diabetes Database (Outcome~$\in\{0,1\}$). To ensure compatibility with the primary pipeline, features were harmonized to the model's input schema. Following common practice for this dataset, physiologically implausible zeros in \textit{Glucose}, \textit{BloodPressure}, \textit{SkinThickness}, \textit{Insulin}, and \textit{BMI} were treated as missing and imputed using the \emph{median values learned from the primary training data}. Continuous variables were then min--max normalized with the \emph{primary training} scaler parameters (frozen). The external set exhibits moderate imbalance (Outcome$=0$: $n{=}500$, $65.1\%$; Outcome$=1$: $n{=}268$, $34.9\%$), see Figure~\ref{fig:pima_distribution}. In line with external validation, no resampling (SMOTE) or re-tuning was performed on PIMA; models and preprocessing were kept fixed from the primary dataset.

\subsection{Decision Rule and Primary Metrics}
Models output class-1 probabilities. We prioritized \emph{threshold-independent} evaluation and therefore did \emph{not} optimize an operating threshold. Performance on internal and external cohorts is primarily summarized with ROC and PR curves and their areas (AUROC, AUPRC), alongside calibration (Brier score, calibration slope/intercept, and reliability plots). For SVM, calibrated probabilities were obtained prior to evaluation (Platt scaling on the primary training data, retained in the frozen pipeline).

For completeness, \emph{threshold-dependent} metrics---accuracy, precision, recall, and F1-score values were computed at a fixed default decision rule $\tau = 0.5$; no threshold tuning was performed on the external cohort.

\subsection{External Validation Protocol}
External performance was assessed on the PIMA Indians Diabetes dataset (N = 768) using the \emph{frozen} pipeline learned on the primary cohort. No model parameters, hyperparameters, or preprocessing statistics were refit on the external data, and no resampling was applied.

\subsubsection{Pipeline Freezing and Harmonization}
We fixed (i) encoder mappings for categorical variables, (ii) imputation parameters, (iii) scaling parameters, and (iv) the probability calibration mapping (for SVM via Platt scaling) learned on the primary training split. PIMA features were mapped to the model input schema, implausible zeros were imputed with primary-training medians, and all continuous variables were scaled using the primary-training scaler. \emph{No} operating-threshold selection was carried out on PIMA; when thresholded metrics are reported, we used $\tau = 0.5$.

\subsubsection{Evaluation}
We reported threshold-independent metrics (AUROC and AUC--PR; the PR baseline equals the positive prevalence in the evaluated cohort) and, secondarily, thresholded metrics at the fixed $\tau=0.5$. Calibration was summarized by slope, intercept, and Brier score, with reliability plots for visual assessment. Ninety-five percent confidence intervals were obtained via bootstrap (1{,}000 resamples); differences in AUROC were tested with DeLong's method and differences in accuracy at $\tau=0.5$ with McNemar’s test. Any recalibration (e.g., isotonic or temperature scaling) was explored only as a sensitivity analysis and did not alter the primary external validation results.

\subsection{Machine Learning Models}
For the development of the diabetes diagnosis system, a combination of four fundamental machine learning algorithms—Support Vector Machine (SVM), Logistic Regression, Random Forest, and Extreme Gradient Boosting (XGBoost)—was employed. To compare the performance, these classifiers were hybridized using a Voting Classifier, which integrates their predictive capabilities. This ensemble approach harnesses the individual strengths of each model to optimize classification accuracy and improve generalization ability \cite{ref10}.

\subsubsection{Support Vector Machine}
It is a margin-based classification algorithm that separates data using a maxi-mum-margin hyperplane. It uses supervised learning where the decision boundary is optimized to reduce generalization error on unseen data \cite{ref11}. The classifier identifies the support vectors from training data that are closest to the margin. These vectors influence the orientation and position of the hyperplane. It per-forms well on both linear and non-linear classification tasks using kernel functions \cite{ref12}. The radial basis function (RBF) is commonly used to handle non-linear data by transforming it into higher dimensions. In medical applications, it was applied to predict spontaneous breathing trial out-comes using weak signal inputs \cite{ref13}.

\begin{equation}
f(\mathbf{x}) = \text{sign}\!\left( \sum_{i=1}^{n} \alpha_i y_i \langle \mathbf{x}_i, \mathbf{x} \rangle + b \right)
\label{eq:svm}
\end{equation}

\subsubsection{Logistic Regression}
It is a supervised classification algorithm used to predict binary outcomes. It estimates the probability that a data instance belongs to a particular class using a logistic function \cite{ref11}. It transforms the linear combination of input features into a probability using the sigmoid function. The algorithm assumes no multicollinearity among predictors and linearly separates classes in the log-odds space \cite{ref14}. Its coefficients can be estimated using maximum likelihood estimation, allowing robust statistical inference in high-dimensional datasets \cite{ref15}.

\begin{equation}
P(y=1 \mid \mathbf{x}) = \frac{1}{1 + \exp\!\left(-(\beta_0 + \sum_{j=1}^{p} \beta_j x_j)\right)}
\label{eq:logreg}
\end{equation}

\subsubsection{Random Forest}
It is an ensemble-based classification technique that operates by constructing multiple decision trees. Each tree is trained on a bootstrapped subset of data using a random subset of features \cite{ref16}. The output of the forest is determined by aggregating predictions from individual trees through majority voting. It helps reduce overfitting, which is common in single decision trees. The method is robust to noise and performs well with both categorical and numerical data \cite{ref17}. It can rank feature importance, making it valuable in feature selection. In medical prediction tasks, it has shown strong performance in disease diagnosis using structured clinical indicators \cite{ref18}.

\begin{equation}
\hat{y} = \text{majority\_vote}\!\left( h_1(\mathbf{x}), h_2(\mathbf{x}), \ldots, h_T(\mathbf{x}) \right)
\label{eq:rf}
\end{equation}

\subsubsection{Extreme Gradient Boosting}
It is an optimized gradient boosting algorithm designed for high performance and scalability. XGBoost builds decision trees sequentially, where each new tree corrects the errors made by previous ones using gradient descent \cite{ref11}. It introduces regularization to avoid overfitting and handles missing data efficiently. It supports parallelization, making it faster than traditional boosting methods. In healthcare applications, it has been employed to predict mortality and surgical complications due to its robustness \cite{ref19}. It also includes built-in cross-validation and handles high-dimensional sparse datasets \cite{ref14}.

\begin{equation}
\hat{y}_i^{(t)} = \hat{y}_i^{(t-1)} + \eta f_t(\mathbf{x}_i), 
\qquad \mathcal{L}^{(t)} = \sum_{i=1}^{n} l\!\left(y_i, \hat{y}_i^{(t)}\right) + \sum_{t=1}^{T} \Omega(f_t)
\label{eq:xgboost}
\end{equation}

\subsubsection{Voting Classifier Algorithm}
Voting Classifier is a meta-model that integrates multiple base classifiers to form a robust ensemble learning system. It operates using two main strategies: hard voting, where each model votes for a class and the majority wins, and soft voting, where predicted probabilities from all models are averaged to select the class with the highest probability \cite{ref20,ref21}. This ensemble technique benefits from the diversity of the base models and reduces overfitting by combining their predictions \cite{ref22}. It does not depend on the internal mechanics of the classifiers and can blend heterogeneous models, making it highly flexible \cite{ref23}. Voting Classifier also supports weighted voting, allowing greater influence on models with higher individual performance \cite{ref24}.

\begin{equation}
\hat{y} = \arg\max_{c \in \mathcal{C}} \sum_{m=1}^{M} w_m \cdot \mathbf{1}\!\left(h_m(\mathbf{x}) = c\right)
\label{eq:voting}
\end{equation}

\subsubsection{Performance Metrics}
Performance evaluation in machine learning relies on robust metrics to capture predictive accuracy and generalization ability. Accuracy provides the ratio of correct predictions but may be misleading in imbalanced datasets \cite{ref25}. Precision quantifies the proportion of relevant positive predictions, while recall captures the ability to detect all relevant instances \cite{ref26,ref27}. The F1-score, a harmonic mean of precision and recall, balances both metrics, especially when trade-offs exist \cite{ref28}. AUC-PR (Area Under the Precision-Recall Curve) is more informative than AUC-ROC in skewed class distributions, offering insight into model confidence for minority class detection \cite{ref29,ref30}. Together, these metrics form a comprehensive framework for evaluating classifier performance across various domains.

\section{Results}\label{sec:sec4}

\subsubsection{Primary Dataset (threshold-independent evaluation)}
\begin{figure}[ht]
  \centering
  \begin{minipage}{0.48\linewidth}
    \centering
    \includegraphics[width=\linewidth]{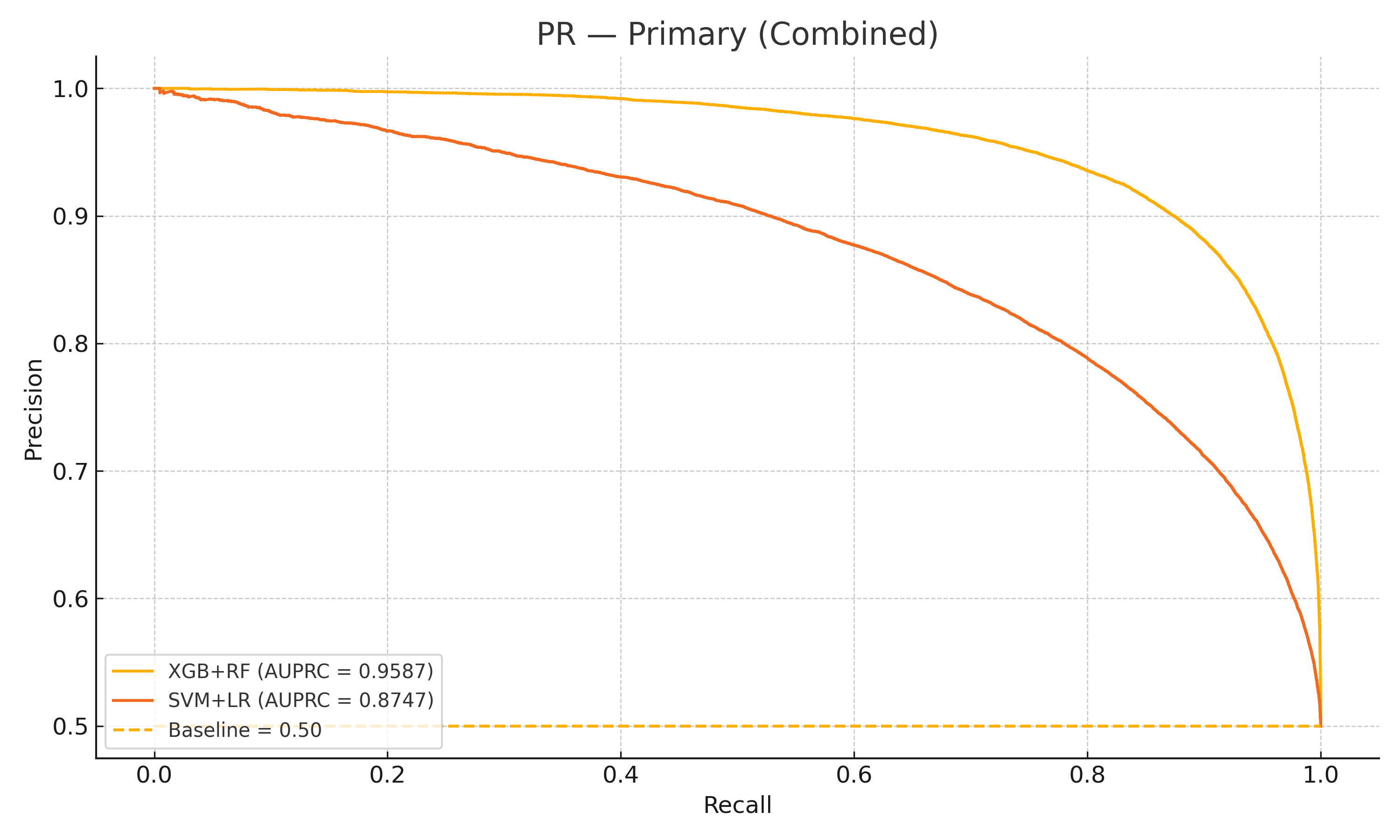}
    \subcaption{Primary dataset}\label{fig:pr_primary_combined}
  \end{minipage}\hfill
  \begin{minipage}{0.48\linewidth}
    \centering
    \includegraphics[width=\linewidth]{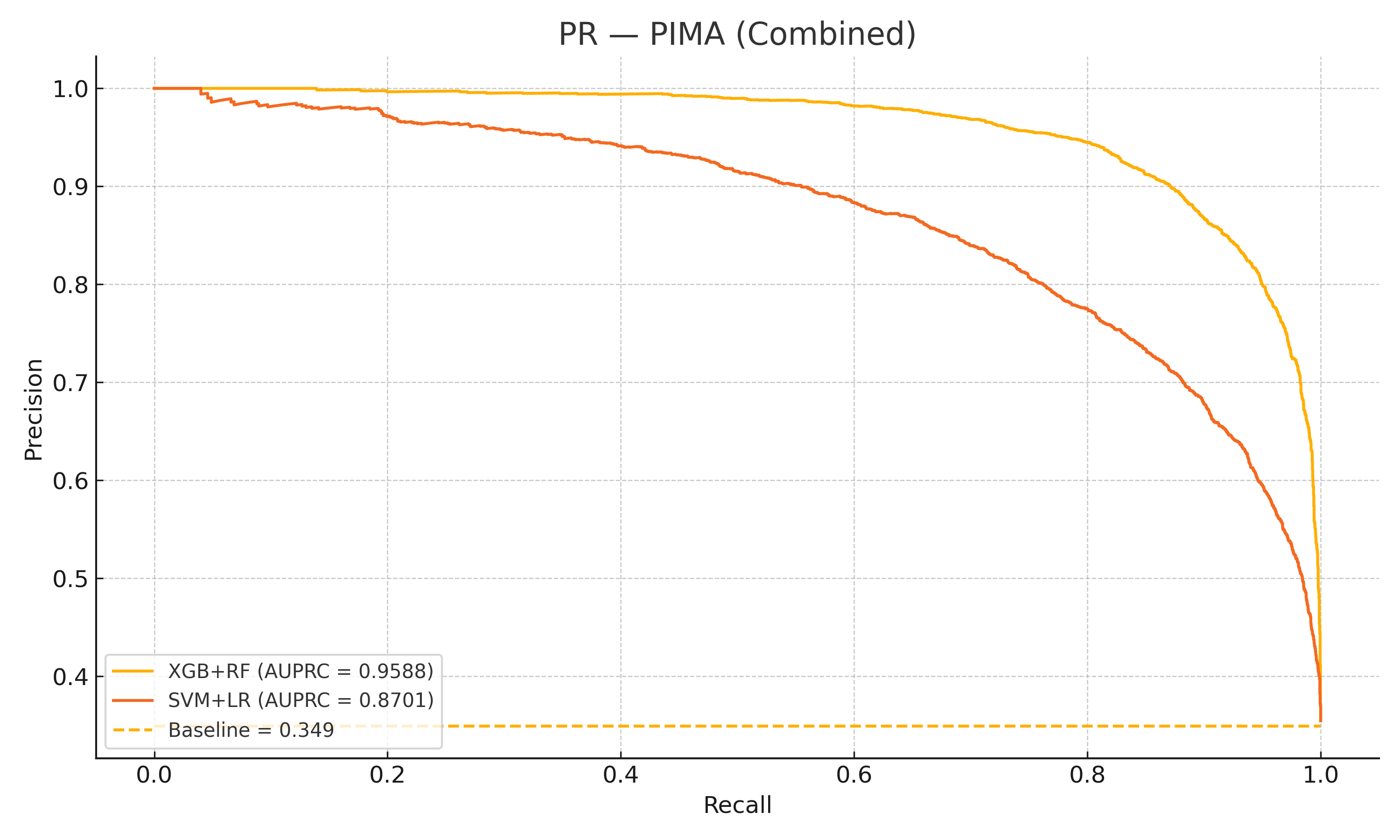}
    \subcaption{External (PIMA) dataset}\label{fig:pr_pima_combined}
  \end{minipage}
  \caption{Precision–Recall curves for the primary and external datasets.}
  \label{fig:curve_both}
\end{figure}

On the primary cohort, threshold-independent metrics indicate excellent discrimination for both hybrids, with XGBoost--RF clearly dominant. The combined PR and ROC curves are shown in Figure~\ref{fig:pr_primary_combined} and Figure~\ref{fig:roc_primary_combined}, respectively. The area under the PR curve (AUPRC; baseline $=0.50$ because the evaluation split is balanced) was highest for \textit{XGB--RF} ($\mathrm{AUPRC}\approx0.998$) and remained high for \textit{SVM--LR} ($\mathrm{AUPRC}\approx0.947$). The corresponding ROC areas were $\mathrm{AUROC}\approx0.995$ for \textit{XGB--RF} and $\mathrm{AUROC}\approx0.978$ for \textit{SVM--LR}.

\begin{table}[ht]
\centering
\small
\setlength{\tabcolsep}{5pt} 
\begin{threeparttable}
\caption{Performance on primary and external (PIMA) diabetes datasets}
\label{tab:perf_primary_external}
\begin{tabularx}{\linewidth}{@{}lYYYY@{}}
\toprule
 & \multicolumn{2}{c}{\textbf{Primary}} & \multicolumn{2}{c}{\textbf{External (PIMA)}} \\
\cmidrule(lr){2-3}\cmidrule(lr){4-5}
\textbf{Metric} & \textbf{XGB--RF} & \textbf{SVM--LR} & \textbf{XGB--RF} & \textbf{SVM--LR} \\
\midrule
Accuracy  & 0.997 & 0.920 & 0.960 & 0.900 \\
Precision & 0.998 & 0.920 & 0.941 & 0.855 \\
Recall    & 0.995 & 0.920 & 0.944 & 0.858 \\
F1-score  & 0.997 & 0.920 & 0.942 & 0.857 \\
\bottomrule
\end{tabularx}
\end{threeparttable}
\end{table}

\subsubsection{External Validation on PIMA (frozen pipeline; no tuning)}
\begin{figure}[ht]
  \centering
  \begin{minipage}{0.48\linewidth}
    \centering
    \includegraphics[width=\linewidth]{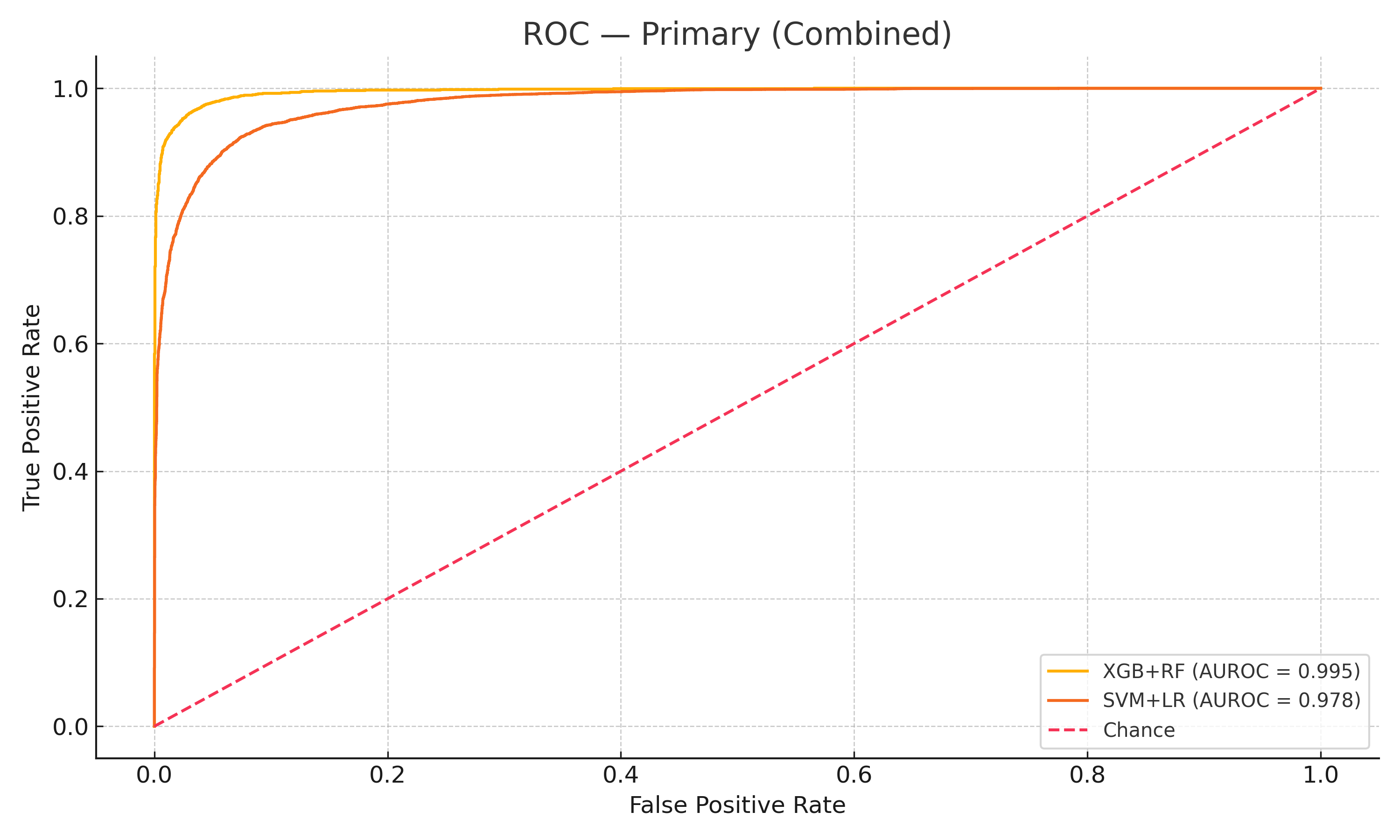}
    \subcaption{Primary dataset}\label{fig:roc_primary_combined}
  \end{minipage}\hfill
  \begin{minipage}{0.48\linewidth}
    \centering
    \includegraphics[width=\linewidth]{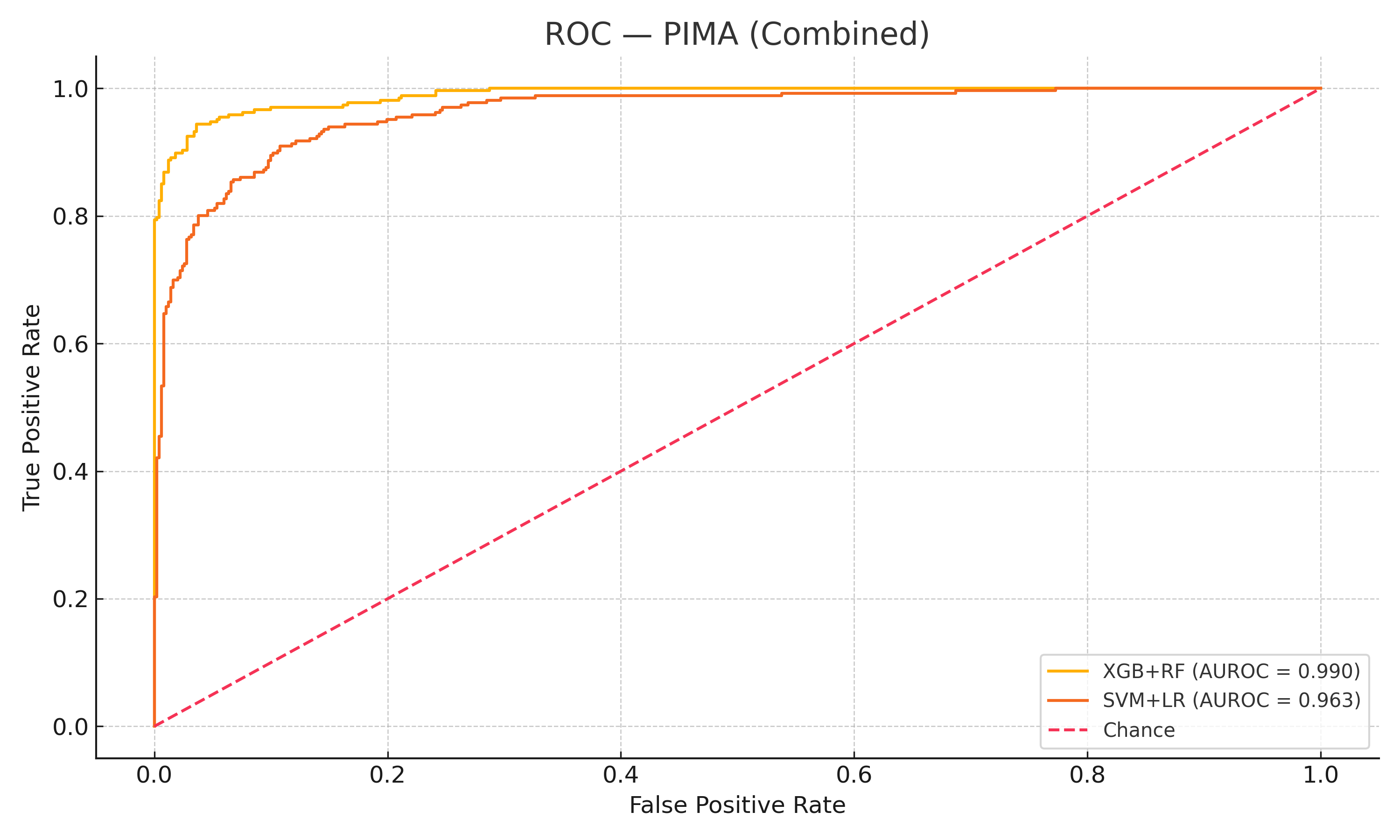}
    \subcaption{External (PIMA) dataset}\label{fig:roc_pima_combined}
  \end{minipage}
  \caption{ROC curves for the primary and external datasets.}
  \label{fig:curve_both}
\end{figure}

Applying the frozen preprocessing and models to the PIMA cohort yielded high and clinically meaningful discrimination for both hybrids (Figures~\ref{fig:pr_pima_combined} and~\ref{fig:roc_pima_combined}). With a PR baseline equal to the positive prevalence in PIMA ($\approx0.349$), \textit{XGB--RF} achieved $\mathrm{AUPRC}\approx0.959$ and \textit{SVM--LR} $\mathrm{AUPRC}\approx0.875$, both well above baseline. ROC areas were likewise high ($\mathrm{AUROC}\approx0.990$ for \textit{XGB--RF} and $\approx0.963$ for \textit{SVM--LR}). 

\subsubsection{Comparative Analysis and Generalization.}
The within-dataset ranking was consistent across cohorts: Hybrid XGBoost--RF
outperformed Hybrid SVM--LR on every metric in both the internal and external
evaluations. Absolute performance decreased on PIMA for both models, as expected under distributional and case-mix shift, but the magnitude of shrinkage was modest for XGBoost--RF (changes from primary to PIMA:
$\Delta$Accuracy $=-0.037$, $\Delta$Precision $=-0.057$, $\Delta$Recall $=-0.051$, $\Delta$F1 $=-0.055$, $\Delta$AUC\textendash PR $\approx -0.040$) and larger for SVM--LR ($\Delta$Accuracy $=-0.020$, $\Delta$Precision $=-0.065$, $\Delta$Recall $=-0.062$, $\Delta$F1 $=-0.063$, $\Delta$AUC\textendash PR $\approx -0.072$). Notably, the between-model gap in AUC\textendash PR widened from $0.051$ on the primary dataset to $0.084$ on PIMA, suggesting that XGBoost--RF retained more transferable signal under cohort shift. Overall, these findings indicate that while both hybrids are effective, Hybrid XGBoost--RF exhibits superior external validity and is the preferred candidate for deployment given its more stable generalization profile.

\section{Conclusion}\label{sec:sec5}
We presented a comparative study of two hybrid ensembles for early diabetes risk stratification---\emph{XGBoost--Random Forest (XGB--RF)} and \emph{SVM--Logistic Regression (SVM--LR)}---evaluated under a \emph{frozen} preprocessing pipeline and a methodology that prioritizes \emph{threshold-independent} validation (ROC/PR) with calibration. All models produced class-1 probabilities; no operating thresholds were tuned on the external cohort. Any thresholded metrics reported for PIMA were computed at the default rule \(\tau=0.5\) and are secondary to ROC/PR.

On the \textbf{primary} cohort, both hybrids discriminated well, with \textit{XGB--RF} showing near-perfect curves (AUROC \(\approx 0.995\); AUPRC \(\approx 0.998\), baseline \(=0.50\)) and \textit{SVM--LR} performing strongly though noticeably lower (AUROC \(\approx 0.978\); AUPRC \(\approx 0.947\)). The predicted-probability distributions reflected clear separation (higher medians for positives; lower medians for negatives) and good calibration, consistent with the PR/ROC envelopes (cf.\ Figures~\ref{fig:pr_primary_combined}, \ref{fig:roc_primary_combined}). For completeness, thresholded metrics at \(\tau=0.5\) (Table~\ref{tab:perf_primary_external}) aligned with these rankings.

Under \textbf{external validation} on PIMA---with the frozen encoder, imputer, scaler, and probability calibration applied unchanged---both hybrids retained clinically meaningful discrimination (Figures~\ref{fig:pr_pima_combined}, \ref{fig:roc_pima_combined}). \textit{XGB--RF} achieved AUROC \(\approx 0.990\) and AUPRC \(\approx 0.959\) (baseline \(\approx 0.349\)), while \textit{SVM--LR} reached AUROC \(\approx 0.963\) and AUPRC \(\approx 0.875\). Probability summaries showed that, relative to \textit{SVM--LR}, \textit{XGB--RF} maintained higher central tendencies for positives and lower for negatives, with slightly better calibration, explaining the larger margin above the PR baseline. Thresholded metrics on PIMA (computed at \(\tau=0.5\)) followed the same pattern and are reported only as descriptive context (Table~\ref{tab:perf_primary_external}).

\textbf{Model comparison and generalization.} Across both datasets, \textit{XGB--RF} consistently dominated \textit{SVM--LR} on ROC and PR. Generalization, judged by threshold-independent metrics, showed modest attenuation for \textit{XGB--RF} (AUPRC \(\downarrow\) from \(\approx 0.998\) to \(\approx 0.959\); AUROC \(\approx 0.995 \to 0.990\)) and a larger drop for \textit{SVM--LR} (AUPRC \(\approx 0.947 \to 0.875\); AUROC \(\approx 0.978 \to 0.963\)). The widening gap in AUPRC on PIMA indicates that the gradient-boosting + bagging hybrid preserves ranking information more robustly under cohort and prevalence shift.

\textbf{Implications.} From a health-informatics standpoint, both hybrids are viable decision-support components for proactive diabetes screening, but \textit{XGB--RF} exhibits superior external validity and more stable probability estimates, making it the preferred candidate for deployment in web-based or clinical workflows. Because we foreground threshold-independent evidence, operational thresholds for specific clinical use cases can later be selected (e.g., via decision-curve or cost-sensitive analysis) without re-tuning the model on external data.

In summary, under a leakage-safe frozen pipeline and threshold-independent evaluation, \textit{XGB--RF} generalizes better than \textit{SVM--LR} and is the stronger choice for reliable, transferable diabetes risk stratification.

\section{Future Work}\label{sec:sec6}
Future work should evaluate advanced hybridization that couples strong tabular learners (e.g., gradient boosting) with modern neural backbones (transformers for longitudinal signals, graph neural networks for comorbidity structure), supported by self-supervised pretraining on large unlabeled EHR corpora and automated model selection (Bayesian optimization/NAS) with sparsity-inducing feature selection. Robustness and trustworthiness warrant domain generalization or test-time adaptation for dataset shift, rigorous calibration and uncertainty quantification (temperature/isotonic scaling, deep ensembles, conformal prediction), and fairness auditing with mitigation. Privacy-preserving, multi-site development via federated learning (with secure aggregation and, where appropriate, differential privacy) and multimodal fusion of structured data, clinical text, and imaging should be assessed for incremental utility. Clinical translation should include longitudinal, multi-institution external validation, prospective workflow studies, cost-effectiveness analyses across triage thresholds, and post-deployment monitoring (drift detection, periodic recalibration, safe continual learning), adhering to TRIPOD-AI and related reporting standards.

\section*{Funding}
 This research did not receive any specific grant from funding agencies in the public, commercial, or not-for-profit sectors.

\section*{Data Availability}
All data analyzed in this study are \emph{publicly available} from third-party repositories; no new patient data were collected by the authors.

\paragraph{Primary diabetes dataset.}
The “Diabetes Prediction” dataset (100{,}000 records; eight risk variables plus a binary diabetes label) is available on Kaggle at
\url{https://www.kaggle.com/datasets/iammustafatz/diabetes-prediction-dataset}.
Access and reuse are subject to Kaggle’s Terms of Service.

\paragraph{External validation dataset (PIMA).}
The PIMA Indians Diabetes Database (768 records; eight predictors and a binary outcome) is available via the UCI/Kaggle mirror at
\url{https://www.kaggle.com/datasets/uciml/pima-indians-diabetes-database?resource=download}. Access and reuse are subject to Kaggle’s Terms of Service and the original dataset license.

\paragraph{Derived data.}
All derived artifacts (e.g., train/validation/test splits, frozen encoder/scaler parameters, and model outputs) can be reproduced from the procedures described in the \emph{Materials and Methods} section using the publicly available datasets above. No restrictions apply beyond those imposed by the hosting repositories.

\section*{Ethics Approval and Informed Consent}
This study involved secondary analysis of publicly available, de-identified datasets (the “Diabetes Prediction” dataset and the PIMA Indians Diabetes Database; see Data Availability). No new data were collected, and no identifiable private information was accessed. In accordance with our institution’s policy, this work was determined to be \emph{not human subjects research} / \emph{exempt from review} (no informed consent
required). Data use complied with the terms and licenses of the hosting repositories. No experiments on humans or animals were performed by the authors.

\section*{Declaration of Competing Interest}
The authors declare no competing interests.

\end{document}